\documentclass{article} 
\usepackage{iclr2015,times}
\usepackage{hyperref}
\usepackage{url}
\usepackage{graphicx}
\usepackage{psgo} 
\graphicspath{{fig/}}
\usepackage{textcomp}

\title{Move Evaluation in Go Using Deep \\Convolutional Neural Networks}

\iclrconference
\iclrfinalcopy

\author{
Chris J. Maddison\\
University of Toronto\\
\texttt{cmaddis@cs.toronto.edu}\\
\And
Aja Huang$^{1}$, Ilya Sutskever$^{2}$, David Silver$^{1}$ \\
Google DeepMind$^1$, Google Brain$^2$ \\
\texttt{\{ajahuang,ilyasu,davidsilver\}@google.com} \\
}

\begin{document}
\maketitle

\begin{abstract}

The game of Go is more challenging than other board games, due to the difficulty of constructing a position or move evaluation function. In this paper we investigate whether deep convolutional networks can be used to directly represent and learn this knowledge. We train a large 12-layer convolutional neural network by supervised learning from a database of human professional games. The network correctly predicts the expert move in 55\% of positions, equalling the accuracy of a 6 dan human player. When the trained convolutional network was used directly to play games of Go, without any search, it beat the traditional-search program \emph{GnuGo} in 97\% of games, and matched the performance of a state-of-the-art Monte-Carlo tree search that simulates two million positions per move.

\end{abstract}

\section{Introduction}

The most frequently cited reason for the difficulty of Go, compared to games such as Chess, Scrabble or Shogi,
is the difficulty of constructing an evaluation function that can differentiate good moves from bad in a given position. The combination of an enormous state space of $10^{170}$ positions, combined with sharp tactics that lead to steep non-linearities in the optimal value function, has led many researchers to conclude that representing and learning such a function is impossible \citep{muller:ai}. In previous years, the most successful methods have sidestepped this problem altogether using Monte-Carlo search, which dynamically evaluates a position through random sequences of self-play. Such programs have led to strong amateur level performance, but a considerable gap still remains between top professional players and the strongest computer programs. The majority of recent progress has been due to increased quantity and quality of prior knowledge, which is used to bias the search towards more promising states in both the search tree and during rollouts \citep{crazystone, gelly:rave, fuego, baudivs2012pachi, erica:balancing}, and it is widely believed that this knowledge is the major bottleneck towards further progress \citep{limitations}. However, this knowledge again is ultimately compiled into an evaluation function or distribution that expresses a preference over moves. 

In this paper we address these fundamental questions of representation and learning of Go knowledge, by using a deep convolutional neural network (CNN). Although CNNs have previously been applied to the game of Go, with modest success \citep{schraudolph, neurogo, sutskever}, previous architectures have typically been limited to one hidden layer of relatively small size, and have not exploited recent advances in computational power. In this paper we use much deeper and larger CNNs of 12 hidden layers and several billion connections to represent and learn Go knowledge. We find that this increase in depth and size leads to a qualitative jump in performance, suggesting that contrary to previous beliefs, a strong move evaluation function for Go can indeed be represented and learnt by such architectures. 

We focus on a supervised learning setup, in which the network is
trained to predict expert human moves, using a large database of
professional $19\times19$ Go games. The predictive accuracy of the CNN
on a held-out set of positions reaches 55\%, which is a significant improvement over the 35\% and 39\% predictive
accuracy reported for some of the strongest Go programs, and comparable to the performance of the 6 dan author on the same data set.
Furthermore, when the CNN was used to
play games by directly selecting the move recommended by the network
output, without any search, it equalled the performance of
state-of-the-art Monte-Carlo search programs, such as \emph{Pachi} \citep{baudivs2012pachi}, that are given 10,000
rollouts per move (i.e., programs that combine handcrafted or
shallow prior knowledge with a search that simulates two million
positions), and the first strong Monte-Carlo search program \emph{MoGo}  with 100,000 rollouts per move \citep{gelly:combined}. In addition, direct move selection using the CNN beat
\emph{GnuGo} (a traditional search program) in 97\% of games.\footnote{
Since we performed this research, we have learned that \cite{storkey} independently adopted a similar approach using a smaller 8-layer CNN to achieve 44\% move prediction accuracy; and defeated GnuGo in 86\% of games.}  

Finally, we demonstrate that the Go knowledge embodied by the CNN can be effectively combined with Monte-Carlo tree search, by using a \emph{delayed prior knowledge} procedure. In this approach, the CNN is evaluated asynchronously on a GPU, and results are incorporated into the main search procedure once available. Using 100,000 rollouts per move, the overall search defeats the raw CNN in $87\%$ of games.

\section{Prior Work}

Convolutional neural networks have a long history in the game of Go. Schraudolph \cite{schraudolph} trained a simple CNN (exploiting rotational, reflectional, and colour inversion symmetries) to predict final territory, by reinforcement learning from games of self-play. The resulting program beat a simplistic handcrafted program called \emph{Wally}. \emph{NeuroGo} \citep{neurogo} used a more sophisticated architecture to predict final territory, eyes, and connectivity, again exploiting symmetries; and used a connectivity pathfinder to propagate information across weakly connected groups of stones. Enzenberger's program also used reinforcement learning from self-play. When combined with an alpha-beta search, \emph{NeuroGo} equalled the performance of \emph{GnuGo} on $9\times9$ Go, and reached around 13 kyu on $19\times19$ Go. \cite{sutskever} applied convolutional networks to supervised learning of expert moves, but using a small 1 hidden layer CNN; this matched the state-of-the-art prediction performance, achieving 34.6\% accuracy, but this was not sufficient to play Go at any reasonable level.

The most successful current programs in Go are based on Monte-Carlo tree search \citep{kocsis}. The basic algorithm was augmented in \emph{MoGo} to use prior knowledge to bootstrap value estimates in the search tree \citep{gelly:combined}; and to use abstractions over subtrees to accelerate the search \citep{gelly:rave}. The strongest current programs such as \emph{CrazyStone} apply supervised learning to construct a move selection policy; this is then used to bias the exploration during search; a faster policy is also learned that selects moves during rollouts \citep{crazystone}. CrazyStone achieved a 35\% move prediction accuracy by extracting a large database of common patterns from expert games, and combining them into a large linear softmax.

Recent work in image recognition has demonstrated considerable advantages of deep convolutional networks over alternative architectures. \cite{krizhevsky2012imagenet} were the first to achieve a very large performance gain with large and deep convolutional neural networks over traditional computer vision systems. Improved convolutional neural network architectures (primarily in the form of deeper 
networks) \citep{oxford_net} provided another substantial improvement, 
culminating with \cite{google_net}, who reduced the error rate of \cite{krizhevsky2012imagenet}
from 15.3\% top-5 error to 7.0\%. The power and generality of large and deep convolutional neural networks
suggests that they may do well on other ``visual'' domains, such as computer Go.

\section{Data}

The dataset used in this work comes from the KGS Go Server. It consists of sequences of board positions $s_t$ for complete games played between humans of varying rank. Board state information includes the position of all stones on the 19x19 board and the sequence allows one to determine the sequence of moves; a move $a_t$ is encoded as a 1 of 361 indicator for each position on the 19x19 board. We collected 29.4 million board-state next-move pairs $(s_t, a_t)$ corresponding to 160,000 games.

Each position $s_t$ was preprocessed into a set of $19\times19$ feature planes $\phi(s_t)$, that serve as input to the neural network. The features that we use come directly from the raw representation of the game rules (stones, liberties, captures, legality, turns since). In addition, we have one simple tactical feature representing a basic common pattern in Go known as ladders; in practice this adds a small performance benefit, but the results that we report would be qualitatively similar even without these features. Many of the features are split into multiple planes of binary values, for example in the case of liberties there are separate binary features representing whether each intersection has 1 liberty, 2 liberties, 3 liberties, $>=4$ liberties. The feature planes are listed in Table \ref{table:features}.\footnote{Due to the computational cost of running extensive experiments, it is possible that some of these features are unnecessary or redundant. }

\begin{table}
\begin{tabular}{lrl}
Feature & Planes & Description \\
\hline
Black / white / empty & 3 & Stone colour \\
Liberties & 4 & Number of liberties (empty adjacent points) \\
Liberties after move & 6 & Number of liberties after this move is played \\
Legality & 1 & Whether point is legal for current player \\
Turns since & 5 & How many turns since a move was played \\
Capture size & 7 & How many opponent stones would be captured \\
Ladder move & 1 & Whether a move at this point is a successful ladder capture \\
KGS rank & 9 & Rank of current player \\
\end{tabular}
\caption{\label{table:features} Features used as inputs to the CNN.}
\end{table}

Finally, we used the following minor innovation.  Our dataset consists of games
from players of different strengths.  Specifically, the KGS data contains more games by lower dan players, and fewer games by higher dan players. As a result, a naive approach to training on the KGS data will result in a network that primarily imitates weaker players. Alternatively, training only on games by stronger players would result in a massive reduction of training data.

To mitigate this, we provided the network with an additional global inputs indicating the
player's rank. Specifically we add 9 feature planes each indicating a specific rank. This is like a 1 of 9 encoding that represents the strength of the current player. That is, if the network is learning to predict a move made by a $d$ dan
player, the $d$th rank feature plane is filled with 1s and the remaining 8 planes are filled with 0s. This has the effect of providing a dynamic bias to the network that depends on rank.

Because every Go game is symmetric under reflections and rotations, we augmented the dataset by sampling uniformly from one of the 8 symmetric boards as we filled minibatches in gradient descent. The dataset was split into a training set of 27.4 million board-state next-move pairs  and a test set of 2 million. This split was done before shuffling, so this corresponds to a test set with distinct games.

\section{Architecture \& Training}

In this section we describe the precise network architecture and the details
of the training procedure. 

We used a deep convolutional neural network with 12 weight matrices for each of 12 layers and rectified linear non-linearities. The first hidden layer's filters were of size 5$\times$5
and the remainder were of size 3$\times$3, with a stride of 1. 
Every layer operated on a $19\times19$ input space, with no pooling; outputs were zero-padded back up up to $19\times19$.  The number of filters in each layer ranged from 64 to 192. In addition to convolutions,
we also used \emph{position-dependent biases} (following \cite{sutskever}).
Our best model has 2.3 million parameters, 630 million connections, and 550,000 hidden units.

The output layer of the CNN was also convolutional with position dependent biases, but with only two filters. Each produced a $19\times19$ plane, corresponding to inputs to two softmax distributions of size 361.  The first 
softmax is the distribution over the next move if it is the black player's
turn, and the second softmax is the distribution over the next move if it is the 
white player's move. Although both players may often prefer the same move, in general the optimal policy may select different moves for each player.

We also experimented with
weight symmetries \cite{schraudolph}. Given that the board is symmetric, it makes sense to force the filters and biases to be rotationally and reflectionally symmetric, by aggregating weight updates over the 8-fold symmetry group between connections. This type of symmetry is stronger than the symmetric data augmentation described above, since it enforces local symmetry of all filters at all locations on the board, not just global symmetry of the entire board. 

For training the network, we used asynchronous stochastic gradient descent \citep{distbelief}
with 50 replicas each on its own GPU.  All parameters were initialized randomly from a uniform[-0.05, 0.05]. Each replica was trained for 25 epochs with a batchsize of 128, a fixed learning rate of 0.128 normalized by batchsize, and no momentum. The network was then fine-tuned on a single GPU with vanilla SGD for 3 epochs with an annealed learning rate, beginning at half the learning rate for the asynchronous setting and halved again every epoch. 
After augmenting the dataset with random symmetries overfitting was very minor --- our 10 layer network overfit by under 1\% achieving 55\% on the training set and 54.5\% on the test set. Even at the end of training errors on the test set did not increase. This suggests that we are currently operating in an underfitting regime suggesting that further improvement is possible. 
All reported accuracies are on a held out test set.

\section{Results}

\subsection{Investigation of Weight Symmetries}

We evaluated the effect of weight symmetries on a smaller CNN
with 3 and 6 layers respectively. These networks were trained on a reduced feature set, excluding rank, liberties after move, capture size, ladder move, and only including a history of one move. The results are given in the table below:     

\begin{center}
\begin{tabular}{lcc}
model & \% Accuracy \\
\hline
3 layer, 64 filters & 43.3 \\
3 layer, 64 filters, symmetric & 44.3\\
6 layer, 192 filters & 49.6\\
6 layer, 192 filters, symmetric & 49.4\\
\end{tabular}
\end{center}

These results suggest that, perhaps surprisingly, weight symmetries have a strong effect on move prediction
for small and shallow networks, but the effect appeared to disappear completely in larger and deeper 
networks.

\subsection{Accuracy and Playing Strength}

To understand how the performance depends on network depth, we trained several networks of different depths. Each CNN used the same architecture as described above, except that the number of $3\times3$ layers was restricted to 3, 6, 10 and 12 respectively. We measured the prediction accuracy on the test set, and also the playing strength of the CNN when it was used to directly select moves. This was achieved by inputting the current position into the network, and selecting the action with maximum probability in the softmax output for the current player. Unless otherwise specified the \emph{KGS rank} feature was set to its maximum setting.

Performance was evaluated against the benchmark program \emph{GnuGo} 3.8, running at its highest level 10. Comparisons are given with reported values for the 3 dan Monte-Carlo search program \emph{Aya}\footnote{http://computer-go.org/pipermail/computer-go/2014-December/007018.html}; simultaneously published results on a somewhat shallower CNN \cite{storkey}\footnote{It should be noted that  \cite{storkey} did not use the highly-predictive \emph{turns since} feature, because they
believed that it would hurt the network's play. This is an interesting hypothesis, which this work does not address.}; and also with the prediction accuracy of a 6 dan human (the second author) on randomly sampled positions from the test set. All games were scored using Chinese rules, refereed by \emph{GnuGo}; duplicate games were excluded from results.

It is apparent from the results that larger and deeper networks have qualitatively better performance than shallow networks, reaching 97\% winning rate against GnuGo for a large 12-layer network compared to 3.4\% for a small 3-layer network. Furthermore, the accuracy on the supervised learning task is clearly strongly correlated with playing performance, demonstrating that the knowledge learnt by the network generalises effectively to the real task of evaluating moves.

\begin{center}
\begin{tabular}{lrrrr}
Depth & Size & \% Accuracy & \% Wins vs. \emph{GnuGo} & stderr \\
\hline
3 layer & 16 filters & 37.5 & 3.4 & $\pm$ 1.1 \\
3 layer &128 filters & 48.0  & 61.8 & $\pm$ 2.6 \\
6 layer & 128 filters & 51.2  & 84.4 & $\pm$ 1.9 \\
10 layer &128 filters & 54.5 & 94.7 & $\pm$ 1.2 \\
12 layer & 128 filters & 55.2 & 97.2 & $\pm$ 0.9 \\
8 layer \citep{storkey}$^4$ & $\leq$ 64 filters & 44.4 & 86 & $\pm$ 2.5 \\
\hline
\emph{Aya} 2014 && 38.8 & 6 & $\pm$ 1.0 \\
\emph{Human} 6 dan && 52 $\pm 5.8$ & 100 \\
\end{tabular}
\end{center}

It is also valuable to know that the correct move is 
within the network's $n$ most confident predictions. If $n$ can be kept small, then this knowledge can be used to
reduce the program's effective search space.  We find that the top-$n$ performance
of our network is quite strong;  in particular, the network is able to predict  
the correct expert move 94\% of the time when $n=10$.  

\begin{figure}[t!!]
\centering
\includegraphics[scale=0.5]{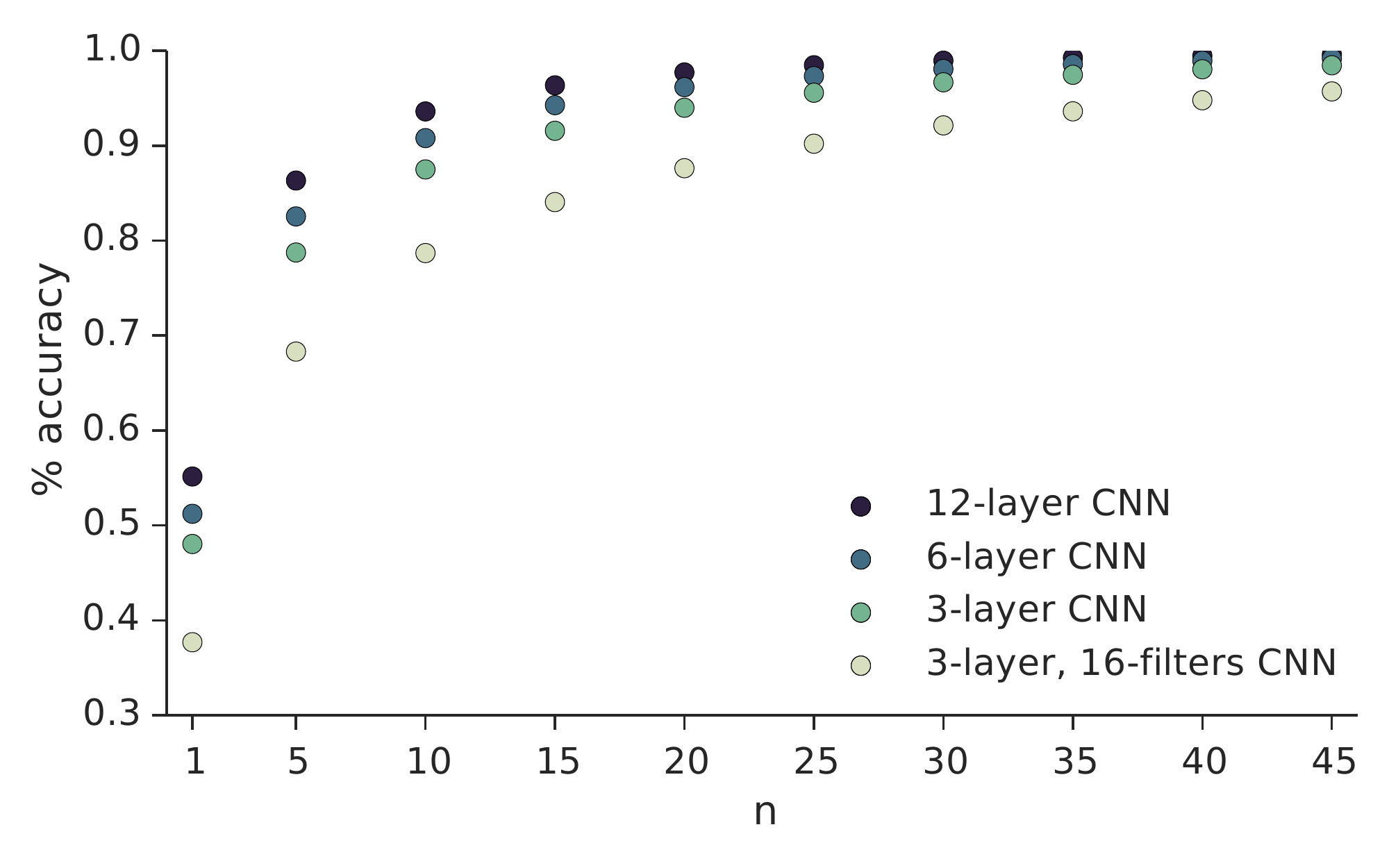}
\caption{\small Probability that the expert's move is within the top-$n$ predictions of the network. The 10 layer CNN was omitted for clarity, but it's performance is only slightly worse than 12 layer. Note $y$-axis begins at 0.30.}
\label{fig:topn}
\vspace{-\baselineskip}
\end{figure}

Next, we compared how the CNN performed when asked to imitate players of different strengths. We used the same CNN, trained on KGS data of all ranks, and asked it to select moves as if it was playing according to a specified rank. The opponent was a fixed 10 layer, 128 filter CNN trained without the \emph{KGS rank} feature. The results clearly show that the network plays significantly better when it is asked to imitate a stronger player. 
\begin{center}
\begin{tabular}{lrr}
KGS rank & \% wins vs. \emph{10-layer CNN} & stderr \\
\hline
1 dan & 49.2 & $\pm$ 3.6 \\
5 dan & 60.1 & $\pm$ 1.6 \\
9 dan & 67.9 & $\pm$ 5.0 \\
\end{tabular}
\end{center}

Finally, we evaluated the overall strength of the 12-layer CNN when used for move selection, by playing against several publicly available benchmark programs. All programs were played at the strongest available settings, and a fixed number of rollouts per move, as specified in the table.

\begin{center}
\begin{tabular}{lrrr}
Opponent & Rollouts per move & Games won by CNN & stderr\\
\hline
\emph{GnuGo} && 97.2 & $\pm$ 0.9 \\
\emph{MoGo} & 100,000 & 45.9 & $\pm$ 4.5 \\
\emph{Pachi} & 100,000 & 11.0 & $\pm$ 2.1 \\
\emph{Fuego} & 100,000 & 12.5 & $\pm$ 5.8 \\ 
\emph{Pachi} & 10,000 & 47.4 & $\pm$ 3.7 \\
\emph{Fuego} & 10,000 & 23.3 & $\pm$ 7.8 \\ 
\end{tabular}
\end{center}

The neural network is considerably stronger than the traditional search-based program GnuGo, and its performance is on a par with \emph{MoGo} with 100,000 rollouts per move \citep{gelly:combined}, and \emph{Pachi} running a somewhat reduced search of 10,000 rollouts per move (a search that visits approximately 2 million positions). It wins more than 10\% of games against \emph{Fuego} (latest svn revision 1966) \citep{fuego} and \emph{Pachi} 10.99 playing at a strong level (using 100,000 rollouts per move over 16 threads).\footnote{The 8-layer network of \cite{storkey} won 12\% of games against the older version Fuego 1.1 at 10 seconds per move on 2 $\times$ 1.6 GHz cores. We tested our 12-layer CNN against Fuego 1.1 at 5 and 10 seconds per move on 2 $\times$ 3.1GHz cores, winning 56\% and 33\% respectively.}

\section{Search} 

The overarching goal of this work is to build a strong Go playing program.  To this
end, we attempted to integrate our move prediction network with Monte Carlo Tree Search (MCTS).  

Combining MCTS with a large deep neural network is far from trivial, since
the CNN is slower than the natural speed of the search, and it is not feasible to evaluate every node with the neural network. The 12-layer network takes 0.15s to evaluate a minibatch of size 128.\footnote{Reducing the minibatch size does not significantly speed up end-to-end computation time in our GPU implementation.}

We address this problem by using \emph{asynchronous} node evaluation.  
In asynchronous node evaluation, MCTS builds its search tree and tracks the new nodes that are added into the search tree. When the number of new nodes equals the minibatch size, all these new positions are submitted to the CNN for evaluation on a GPU. The GPU computes the move recommendations, while the search continues in parallel. Once the GPU computation is complete, the prior knowledge in the new nodes is updated to contain move evaluations from the CNN. The network evaluates the nodes in a FIFO order, in order to maximally influence the search tree. By using a single machine with Intel\textregistered~ Xeon\textregistered ~CPU E5-2643 v2 @ 3.50GHz 
and GeForce GTX Titan Black GPU, we are able to maintain a MCTS search at approximately 47,000 rollouts per second, without dropping CNN evaluations. However, it should be noted that the performance of asynchronous node evaluation is significantly less than a fully synchronous and serial implementation, since new information from the search is only utilised after a significant lag (around 0.15s in our case), due to the GPU computation.

In addition, the MCTS engine utilised standard heuristics for computer Go: RAVE \citep{gelly:rave}, a UCT exploration strategy similar to \cite{chaslot:progressive}, and very simple rollouts based solely on $3\times3$ patterns \citep{erica:balancing}.

We measured the performance of the search-based program by playing games between the 12-layer CNN with MCTS, and a baseline 12-layer CNN without any search. Using 100,000 rollouts per move, the search-based program beats the baseline CNN in 87\% of games. 

\begin{center}
\begin{tabular}{lrr}
Rollouts per move & \% wins against baseline & stderr \\
\hline
100,000 & 86.7 & $\pm$ 3.5 \\
10,000 & 67.6 & $\pm$ 2.6 \\
\end{tabular}
\end{center}

\begin{figure}[t!!]
\centering
\includegraphics[scale=0.55]{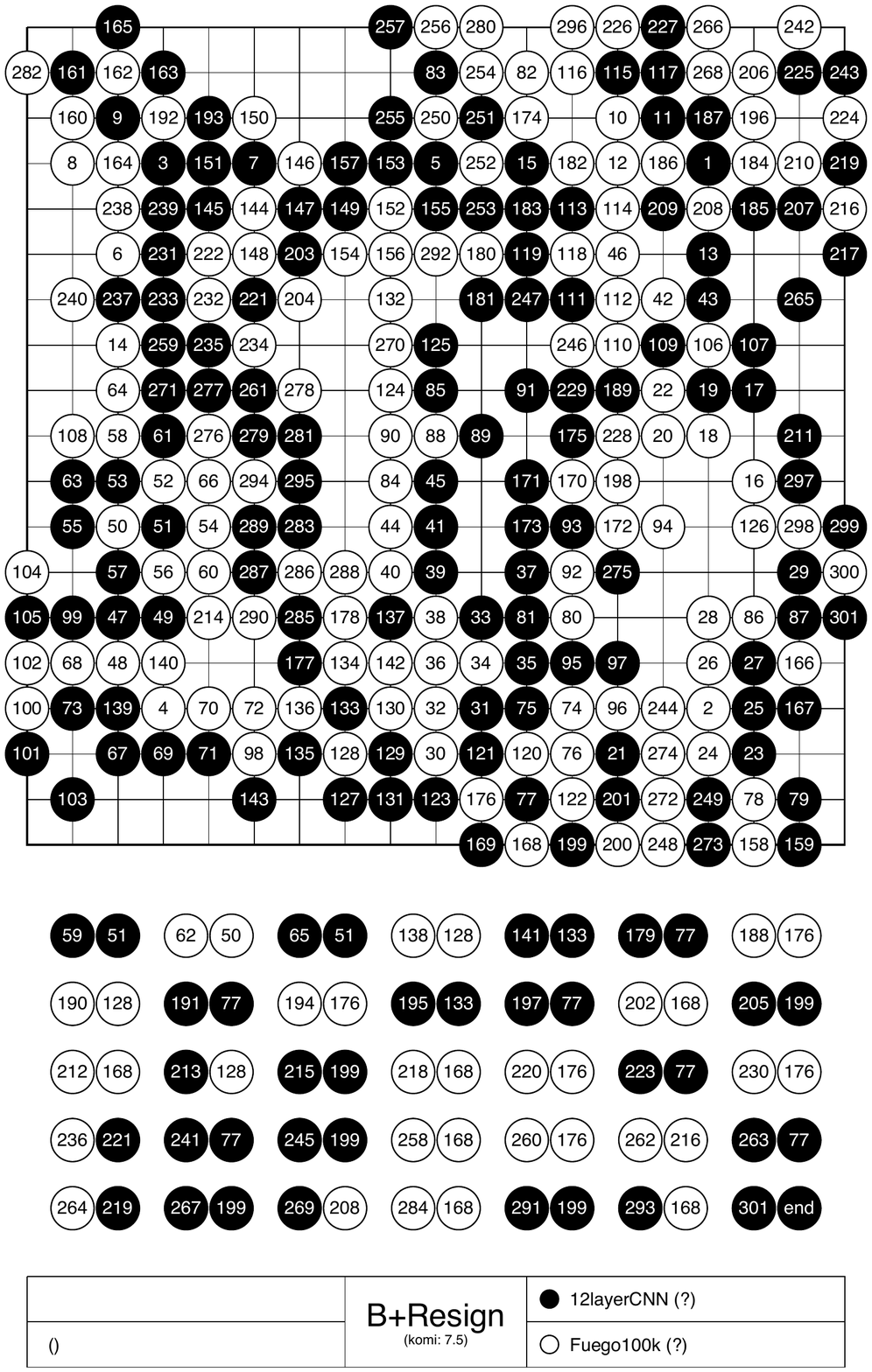}
\caption{\small A game played between the 12-layer CNN (without any search)
and Fuego (using 100k rollouts/move). The CNN plays white.}
\label{fig:game}
\end{figure}

\section{Discussion}

In this work, we showed that large deep convolutional neural networks can predict  
the next move made by Go experts with an accuracy that exceeds previous methods by a large margin, approximately matching human performance. Furthermore, this predictive accuracy translates into much stronger move evaluation and playing strength than has previously been possible. Without any search, the network is able to outperform traditional search based programs such as GnuGo, and compete with state-of-the-art MCTS programs such as \emph{Pachi} and \emph{Fuego}.  

In Figure \ref{fig:game} we present a sample game played by the 12-layer CNN (with no search) versus \emph{Fuego} (searching 100K rollouts per move) which was won by the neural network player. It is clear that the neural network has implicitly understood many sophisticated aspects of Go, including good shape (patterns that maximise long term effectiveness of stones), Fuseki (opening sequences), Joseki (corner patterns), Tesuji (tactical patterns), Ko fights (intricate tactical battles involving repeated recapture of the same stones), territory (ownership of points), and influence (long-term potential for territory). It is remarkable that a single, unified, straightforward architecture can master these elements of the game to such a degree, and without any explicit lookahead. 

On the other hand, we note that the network still has weaknesses: notably it sometimes fails to understand the global picture, behaving as if the life and death status of large groups has been incorrectly assessed. Interestingly, it is precisely these global aspects of the game for which Monte-Carlo search excels, suggesting that these two techniques may be largely complementary. We have provided a preliminary proof-of-concept that MCTS and deep neural networks may be combined effectively. It appears that we now have two core elements that scale effectively with increased computational resource: scalable planning, using Monte-Carlo search; and scalable evaluation functions, using deep neural networks. In the future, as parallel computation units such as GPUs continue to increase in performance, we believe that this trajectory of research will lead to considerably stronger programs than are currently possible.

\bibliographystyle{iclr2015}
\bibliography{deepgo}

\end{document}